\documentclass{article}

\usepackage{microtype}
\usepackage{graphicx}
\usepackage{subfigure}
\usepackage{booktabs} 

\usepackage{hyperref}

\usepackage[accepted]{icml2025}

\usepackage{amsmath}
\usepackage{amssymb}
\usepackage{mathtools}
\usepackage{amsthm}
\usepackage{wrapfig}
\usepackage{listings}
\usepackage{arydshln}
\usepackage{tikz}
\usepackage{mathpazo}

\newcounter{row}
\newcounter{col}

\newcommand\setrow[9]{
  \setcounter{col}{1}
  \foreach \n in {#1, #2, #3, #4, #5, #6, #7, #8, #9} {
    \edef\x{\value{col} - 0.5}
    \edef\y{9.5 - \value{row}}
    \node[anchor=center] at (\x, \y) {\n};
    \stepcounter{col}
  }
  \stepcounter{row}
}

\usepackage[capitalize,noabbrev]{cleveref}

\usepackage{xcolor}

\definecolor{mydarkblue}{rgb}{0.0,0.15,0.7}
\hypersetup{%
colorlinks=true,
linkcolor=mydarkblue,
citecolor=mydarkblue,
filecolor=mydarkblue,
urlcolor=mydarkblue}
\newcommand{\highlightgreen}[1]{\colorbox{green!15}{#1}}

\theoremstyle{plain}

\theoremstyle{definition}

\theoremstyle{remark}

\usepackage[textsize=tiny]{todonotes}

\icmltitlerunning{Recursive Reasoning with Tiny Networks}

\begin{document}

\twocolumn[
\icmltitle{Less is More: Recursive Reasoning with Tiny Networks}



\icmlsetsymbol{equal}{*}

\begin{icmlauthorlist}
\end{icmlauthorlist}

\begin{center}
{\bf Alexia Jolicoeur-Martineau} \\ Samsung SAIL Montréal
 \\ alexia.j@samsung.com
\end{center}



\icmlkeywords{reasoning, recurrent, arc-agi}

\vskip 0.3in
]




\begin{abstract}

Hierarchical Reasoning Model (HRM) is a novel approach using two small neural networks recursing at different frequencies. This biologically inspired method beats Large Language models (LLMs) on hard puzzle tasks such as Sudoku, Maze, and ARC-AGI while trained with small models (27M parameters) on small data ($\sim$ 1000 examples). HRM holds great promise for solving hard problems with small networks, but it is not yet well understood and may be suboptimal. We propose Tiny Recursive Model (TRM), a much simpler recursive reasoning approach that achieves significantly higher generalization than HRM, while using a single tiny network with only 2 layers. With only 7M parameters, TRM obtains 45\% test-accuracy on ARC-AGI-1 and 8\% on ARC-AGI-2, higher than most LLMs (e.g., Deepseek R1, o3-mini, Gemini 2.5 Pro) with less than 0.01\% of the parameters.
\end{abstract}

\section{Introduction}
\label{submission}

While powerful, Large Language models (LLMs) can struggle on hard question-answer problems. Given that they generate their answer auto-regressively, there is a high risk of error since a single incorrect token can render an answer invalid. To improve their reliability, LLMs rely on Chain-of-thoughts (CoT) \citep{wei2022chain} and Test-Time Compute (TTC) \citep{snell2024scaling}. CoTs seek to emulate human reasoning by having the LLM to sample step-by-step reasoning traces prior to giving their answer. Doing so can improve accuracy, but CoT is expensive, requires high-quality reasoning data (which may not be available), and can be brittle since the generated reasoning may be wrong. To further improve reliability, test-time compute can be used by reporting the most common answer out of $K$ or the highest-reward answer \citep{snell2024scaling}. 

\begin{figure}[H]
    \centering
    \includegraphics[width=0.75\linewidth]{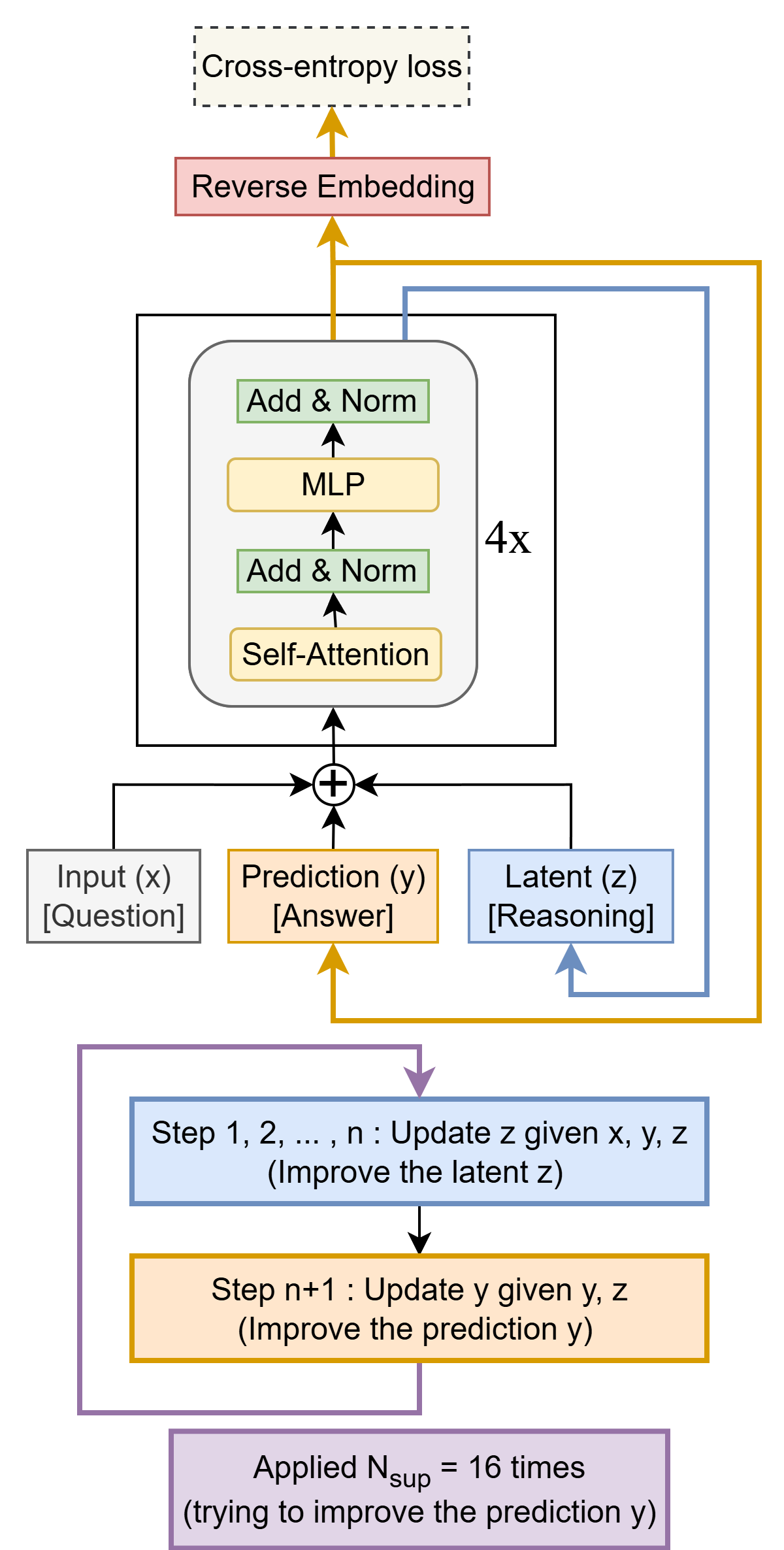}
    \caption{Tiny Recursion Model (TRM) recursively improves its predicted answer $y$ with a tiny network. It starts with the embedded input question $x$ and initial embedded answer $y$, and latent $z$. For up to $N_{sup}=16$ improvements steps, it tries to improve its answer $y$. It does so by i) recursively updating $n$ times its latent $z$ given the question $x$, current answer $y$, and current latent $z$ (recursive reasoning), and then ii) updating its answer $y$ given the current answer $y$ and current latent $z$. This recursive process allows the model to progressively improve its answer (potentially addressing any errors from its previous answer) in an extremely parameter-efficient manner while minimizing overfitting.}
    \label{fig:main}
\end{figure}

\clearpage

However, this may not be enough. LLMs with CoT and TTC are not enough to beat every problem. While LLMs have made significant progress on ARC-AGI \citep{chollet2019measure} since 2019, human-level accuracy still has not been reached (6 years later, as of writing of this paper). Furthermore, LLMs struggle on the newer ARC-AGI-2 (e.g., Gemini 2.5 Pro only obtains 4.9\% test accuracy with a high amount of TTC) \citep{chollet2025arc,arc_prize}.

An alternative direction has recently been proposed by \citet{wang2025hierarchical}. They propose a new way forward through their novel Hierarchical Reasoning Model (HRM), which obtains high accuracy on puzzle tasks where LLMs struggle to make a dent (e.g., Sudoku solving, Maze pathfinding, and ARC-AGI). HRM is a supervised learning model with two main novelties: 1) \emph{recursive hierarchical reasoning}, and 2) \emph{deep supervision}. 

\textbf{Recursive hierarchical reasoning} consists of recursing multiple times through two small networks ($f_L$ at high frequency and $f_H$ at low frequency) to predict the answer. Each network generates a different latent feature: $f_L$ outputs $z_H$ and $f_H$ outputs $z_L$. Both features ($z_L,z_H$) are used as input to the two networks. The authors provide some biological arguments in favor of recursing at different hierarchies based on the different temporal frequencies at which the brains operate and hierarchical processing of sensory inputs. 

\textbf{Deep supervision} consists of improving the answer through multiple supervision steps while carrying the two latent features as initialization for the improvement steps (after detaching them from the computational graph so that their gradients do not propagate). This provide residual connections, which emulates very deep neural networks that are too memory expensive to apply in one forward pass.

An independent analysis on the ARC-AGI benchmark showed that deep supervision seems to be the primary driver of the performance gains \citep{arc_hrm}. Using \emph{deep supervision} doubled accuracy over single-step supervision (going from $19\%$ to $39\%$ accuracy), while \emph{recursive hierarchical reasoning} only slightly improved accuracy over a regular model with a single forward pass (going from $35.7\%$ to $39.0\%$ accuracy). This suggests that reasoning across different supervision steps is worth it, but the recursion done in each supervision step is not particularly important. 

In this work, we show that the benefit from \emph{recursive reasoning} can be massively improved, making it much more than incremental. We propose Tiny Recursive Model (TRM), an improved and simplified approach using a much smaller tiny network with only 2 layers that achieves significantly higher generalization than HRM on a variety of problems. In doing so, we improve the state-of-the-art test accuracy on Sudoku-Extreme from 55\% to 87\%, Maze-Hard from 75\% to 85\%, ARC-AGI-1 from 40\% to 45\%, and ARC-AGI-2 from 5\% to 8\%.

\section{Background}

HRM is described in Algorithm \ref{fig:pseudocode}. We discuss the details of the algorithm further below.

\subsection{Structure and goal}

The focus of HRM is supervised learning. Given an input, produce an output. Both input and output are assumed to have shape $[B,L]$ (when the shape differs, padding tokens can be added), where $B$ is the batch-size and $L$ is the context-length.

HRM contains four learnable components: the input embedding $f_I(\cdot; \theta_I)$, low-level recurrent network $f_L(\cdot; \theta_L)$, high-level recurrent network $f_H(\cdot; \theta_H)$, and the output head $f_O(\cdot; \theta_O)$. Once the input is embedded, the shape becomes $[B,L,D]$ where $D$ is the embedding size. Each network is a 4-layer Transformers architecture \citep{vaswani2017attention}, with RMSNorm \citep{zhang2019root}, no bias \citep{chowdhery2023palm}, rotary embeddings \citep{su2024roformer}, and SwiGLU activation function \citep{hendrycks2016gaussian, shazeer2020glu}.

\begin{figure}[h]
  \centering
    \begin{lstlisting}[language=python]
def hrm(z, x, n=2, T=2): # hierarchical reasoning
    zH, zL = z
    with torch.no_grad():
        for i in range(nT - 2):
            zL = L_net(zL, zH, x)
            if (i + 1) % T == 0:
                zH = H_net(zH, zL)
    # 1-step grad
    zL = L_net(zL, zH, x)
    zH = H_net(zH, zL)
    return (zH, zL), output_head(zH), Q_head(zH)

def ACT_halt(q, y_hat, y_true):
    target_halt = (y_hat == y_true)
    loss = 0.5*binary_cross_entropy(q[0], target_halt)
    return loss

def ACT_continue(q, last_step):
    if last_step:
        target_continue = sigmoid(q[0])
    else:
        target_continue = sigmoid(max(q[0], q[1])))
    loss = 0.5*binary_cross_entropy(q[1], target_continue)
    return loss

# Deep Supervision
for x_input, y_true in train_dataloader:
    z = z_init
    for step in range(N_sup): # deep supervision
        x = input_embedding(x_input)
        z, y_pred, q = hrm(z, x)
        loss = softmax_cross_entropy(y_pred, y_true)
        # Adaptive computational time (ACT) using Q-learning
        loss += ACT_halt(q, y_pred, y_true)
        _, _, q_next = hrm(z, x) # extra forward pass
        loss += ACT_continue(q_next, step == N_sup - 1)
        z = z.detach()
        loss.backward()
        opt.step()
        opt.zero_grad()
        if q[0] > q[1]: # early-stopping
            break
    \end{lstlisting}
  \caption{Pseudocode of Hierarchical Reasoning Models (HRMs).}
  \label{fig:pseudocode}
  \vspace{-5pt}
\end{figure}

\subsection{Recursion at two different frequencies}

Given the hyperparameters used by \citet{wang2025hierarchical} ($n=2$ $f_L$ steps, 1 $f_H$ steps; done $T=2$ times), a forward pass of HRM is done as follows:
\begin{align*}
x &\gets f_I(\tilde{x}) \\
    z_L &\gets f_L\left( z_L + z_H + x \right) \hspace{6pt} \text{\# without gradients} \\
    z_L &\gets f_L\left( z_L + z_H + x \right)\hspace{6pt} \text{\# without gradients} \\
    z_H &\gets f_H\left( z_L + z_H \right)\hspace{22pt} \text{\# without gradients} \\
    z_L &\gets f_L\left( z_L + z_H + x \right)\hspace{6pt} \text{\# without gradients} \\
    z_L &\gets z_L.detach() \\
    z_H &\gets z_H.detach() \\
z_L &\gets f_L\left( z_L + z_H + x \right) \hspace{6pt} \text{\# with gradients} \\
z_H &\gets f_H\left( z_L + z_H \right) \hspace{22pt}\text{\# with gradients} \\
\hat{y} &\gets \text{argmax}(f_O\left( z_H \right))
\end{align*}

where $\hat{y}$ is the predicted output answer, $z_L$ and $z_H$ are either initialized embeddings or the embeddings of the previous deep supervision step (after detaching them from the computational graph). As can be seen, a forward pass of HRM consists of applying 6 function evaluations, where the first 4 function evaluations are detached from the computational graph and are not back-propagated through. The authors uses  $n=2$ with $T=2$ in all experiments, but HRM can be generalized by allowing for an arbitrary number of L steps ($n$) and recursions ($T$) as shown in Algorithm \ref{fig:pseudocode}.

\subsection{Fixed-point recursion with 1-step gradient approximation}

Assuming that ($z_L$, $z_H$) reaches a fixed-point ($z_L^{*}$, $z_H^{*}$) through recursing from both $f_L$ and $f_H$,
\begin{align*}
z_L^{*} &\approx f_L\left( z_L^{*} + z_H + x \right) \\
z_H^{*} &\approx f_H\left( z_L + z_H^{*} \right),
\end{align*}
the Implicit Function Theorem \citep{krantz2002implicit} with the 1-step gradient approximation \citep{bai2019deep} is used to approximate the gradient by back-propagating only the last $f_L$ and $f_H$ steps. 
This theorem is used to justify only tracking the gradients of the last two steps (out of 6), which greatly reduces memory demands.

\subsection{Deep supervision}

To improve effective depth, deep supervision is used. This consists of reusing the previous latent features ($z_H$ and $z_L$) as initialization for the next forward pass. This allows the model to reason over many iterations and improve its latent features ($z_L$ and $z_H$) until it (hopefully) converges to the correct solution. At most $N_{sup}=16$ supervision steps are used.

\subsection{Adaptive computational time (ACT)}

With deep supervision, each mini-batch of data samples must be used for $N_{sup}=16$ supervision steps before moving to the next mini-batch. This is expensive, and there is a balance to be reached between optimizing a few data examples for many supervision steps versus optimizing many data examples with less supervision steps. To reach a better balance, a halting mechanism is incorporated to determine whether the model should terminate early. It is learned through a Q-learning objective that requires passing the $z_H$ through an additional head and running an additional forward pass (to determine if halting now rather than later would have been preferable). They call this method Adaptive computational time (ACT). It is only used during training, while the full $N_{sup}=16$ supervision steps are done at test time to maximize downstream performance. ACT greatly diminishes the time spent per example (on average spending less than 2 steps on the Sudoku-Extreme dataset rather than the full $N_{sup}=16$ steps), allowing more coverage of the dataset given a fixed number of training iterations.

\subsection{Deep supervision and 1-step gradient approximations replaces BPTT}

Deep supervision and the 1-step gradient approximation provide a more biologically plausible and less computationally-expansive alternative to Backpropagation Through Time (BPTT) \citep{werbos1974beyond, rumelhart1985learning, lecun1985procedure} for solving the temporal credit assignment (TCA) \citep{rumelhart1985learning, werbos1988generalization, elman1990finding} problem \citep{lillicrap2019backpropagation}. The implication is that HRM can learn what would normally require an extremely large network without having to back-propagate through its entire depth. Given the hyperparameters used by \citet{jang2023hierarchical} in all their experiments, HRM effectively reasons over $n_{layers}(n+1)TN_{sup}=4*(2+1)*2*16=384$ layers of effective depth.

\subsection{Summary of HRM}

HRM leverages recursion from two networks at different frequencies (high frequency versus low frequency) and deep supervision to learn to improve its answer over multiple supervision steps (with ACT to reduce time spent per data example). This enables the model to imitate extremely large depth without requiring backpropagation through all layers. This approach obtains significantly higher performance on hard question-answer tasks that regular supervised models struggle with. However, this method is quite complicated, relying a bit too heavily on uncertain biological arguments and fixed-point theorems that are not guaranteed to be applicable. In the next section, we discuss those issues and potential targets for improvements in HRM.

\section{Target for improvements in Hierarchical Reasoning Models}

In this section, we identify key targets for improvements in HRM, which will be addressed by our proposed method, Tiny Recursion Models (TRMs).

\subsection{Implicit Function Theorem (IFT) with 1-step gradient approximation} 

HRM only back-propagates through the last 2 of the 6 recursions. The authors justify this by leveraging the Implicit Function Theorem (IFT) and one-step approximation, which states that when a recurrent function converges to a fixed point, backpropagation can be applied in a single step at that equilibrium point. 

There are concerns about applying this theorem to HRM. Most importantly, there is no guarantee that a fixed-point is reached. Deep equilibrium models normally do fixed-point iteration to solve for the fixed point$z^* = f(z^*)$ \citep{bai2019deep}. However, in the case of HRM, it is not iterating to the fixed-point but simply doing forward passes of $f_L$ and $f_H$. To make matters worse, HRM is only doing 4 recursions before stopping to apply the one-step approximation. After its first loop of two $f_L$ and 1 $f_H$ evaluations, it only apply a single $f_L$ evaluation before assuming that a fixed-point is reached for both $z_L$ and $z_H$ ($z_L^* = f_L(z_L^*+z_H+x)$ and $z_H^* = f_H(z_L^*+z_H^*)$). Then, the one-step gradient approximation is applied to both latent variables in succession. 

The authors justify that a fixed-point is reached by depicting an example with $n=7$ and $T=7$ where the forward residuals is reduced over time (Figure 3 in \citet{wang2025hierarchical}). Even in this setting, which is different from the much smaller $n=2$ and $T=2$ used in every experiment of their paper, we observe the following:
\begin{enumerate}
    \item the residual for $z_H$ is clearly well above 0 at every step
    \item the residual for $z_L$ only becomes closer to 0 after many cycles, but it remains significantly above 0
    \item $z_L$ is very far from converged after one $f_L$ evaluation at $T$ cycles, which is when the fixed-point is assumed to be reached and the 1-step gradient approximation is used
\end{enumerate}

Thus, while the application of the IFT theorem and 1-step gradient approximation to HRM has some basis since the residuals do generally reduce over time, a fixed point is unlikely to be reached when the theorem is actually applied.

In the next section, we show that we can bypass the need for the IFT theorem and 1-step gradient approximation, thus bypassing the issue entirely.

\subsection{Twice the forward passes with Adaptive computational time (ACT)}

HRM uses Adaptive computational time (ACT) during training to optimize the time spent of each data sample. Without ACT, $N_{sup}=16$ supervision steps would be spent on the same data sample, which is highly inefficient. They implement ACT through an additional Q-learning objective, which decides when to halt and move to a new data sample rather than keep iterating on the same data. This allows much more efficient use of time especially since the average number of supervision steps during training is quite low with ACT (less than 2 steps on the Sudoku-Extreme dataset as per their reported numbers). 

However, ACT comes at a cost. This cost is not directly shown in the HRM's paper, but it is shown in their official code. The Q-learning objective relies on a halting loss and a continue loss. The continue loss requires an extra forward pass through HRM (with all 6 function evaluations). This means that while ACT optimizes time more efficiently per sample, it requires 2 forward passes per optimization step. The exact formulation is shown in Algorithm \ref{fig:pseudocode}.

In the next section, we show that we can bypass the need for two forward passes in ACT.

\subsection{Hierarchical interpretation based on complex biological arguments}

The HRM's authors justify the two latent variables and two networks operating at different hierarchies based on biological arguments, which are very far from artificial neural networks. They even try to match HRM to actual brain experiments on mice. While interesting, this sort of explanation makes it incredibly hard to parse out why HRM is designed the way it is. Given the lack of ablation table in their paper, the over-reliance on biological arguments and fixed-point theorems (that are not perfectly applicable), it is hard to determine what parts of HRM is helping what and why. Furthermore, it is not clear why they use two latent features rather than other combinations of features.

In the next section, we show that the recursive process can be greatly simplified and understood in a much simpler manner that does not require any biological argument, fixed-point theorem, hierarchical interpretation, nor using two networks. It also explains why 2 is the optimal number of features ($z_L$ and $z_H$).

\begin{figure}[h]
  \centering
    \begin{lstlisting}[language=python]

def latent_recursion(x, y, z, n=6):
    for i in range(n): # latent reasoning
        z = net(x, y, z)
    y = net(y, z) # refine output answer
    return y, z
    
def deep_recursion(x, y, z, n=6, T=3):
    # recursing T-1 times to improve y and z (no gradients needed)
    with torch.no_grad():
        for j in range(T-1):
            y, z = latent_recursion(x, y, z, n)
    # recursing once to improve y and z
    y, z = latent_recursion(x, y, z, n)
    return (y.detach(), z.detach()), output_head(y), Q_head(y)

# Deep Supervision
for x_input, y_true in train_dataloader:
    y, z = y_init, z_init
    for step in range(N_supervision):
        x = input_embedding(x_input)
        (y, z), y_hat, q_hat = deep_recursion(x, y, z)
        loss = softmax_cross_entropy(y_hat, y_true)
        loss += binary_cross_entropy(q_hat, (y_hat == y_true))
        loss.backward()
        opt.step()
        opt.zero_grad()
        if q_hat > 0: # early-stopping
            break
    \end{lstlisting}
  \caption{Pseudocode of Tiny Recursion Models (TRMs).}
  \label{fig:pseudocode2}
\end{figure}

\section{Tiny Recursion Models}

In this section, we present Tiny Recursion Models (TRMs). Contrary to HRM, TRM requires no complex mathematical theorem, hierarchy, nor biological arguments. It generalizes better while requiring only a single tiny network (instead of two medium-size networks) and a single forward pass for the ACT (instead of 2 passes). Our approach is described in Algorithm \ref{fig:pseudocode2} and illustrated in Figure \ref{fig:main}. We also provide an ablation in Table \ref{tab:ablation} on the Sudoku-Extreme dataset (a dataset of difficult Sudokus with only 1K training examples, but 423K test examples). Below, we explain the key components of TRMs. 

\setlength{\tabcolsep}{2pt} 
\begin{table}[h]
    \caption{Ablation of TRM on Sudoku-Extreme comparing \% Test accuracy, effective depth per supervision step $(T(n+1)n_{layers})$, number of Forward Passes (NFP) per optimization step, and number of parameters}
    \centering
    \begin{tabular}{|l|c|c|c|c|}
    \hline
     Method & Acc (\%) & Depth & NFP & \# Params \\
    \hline
HRM & 55.0 & 24 & 2 & 27M \\
\hline
TRM ($T=3, n=6$) & 87.4 & 42 & 1 & 5M \\
w/ ACT & 86.1 & 42 & 2 & 5M \\
w/ separate $f_H, f_L$ & 82.4 & 42 & 1 & 10M \\
no EMA & 79.9 & 42 & 1 & 5M \\
w/ 4-layers, $n=3$  & 79.5 & 48 & 1 & 10M \\
w/ self-attention & 74.7 & 42 & 1 & 7M \\
w/ $T=2, n=2$ & 73.7 & 12 & 1 & 5M \\
w/ 1-step gradient & 56.5 & 42 & 1 & 5M \\
\hline
\end{tabular}
\label{tab:ablation}
\end{table}
\setlength{\tabcolsep}{6pt} 

\subsection{No fixed-point theorem required}

HRM assumes that the recursions converge to a fixed-point for both $z_L$ and $z_H$ in order to leverage the 1-step gradient approximation \citep{bai2019deep}. This allows the authors to justify only back-propagating through the last two function evaluations (1 $f_L$ and 1 $f_H$). To bypass this theoretical requirement, we define a full recursion process as containing $n$ evaluations of $f_L$ and 1 evaluation of $f_H$:
\begin{align*}
z_L &\gets f_L\left( z_L + z_H + x \right) \\
&... \\
z_L &\gets f_L\left( z_L + z_H + x \right) \\
z_H &\gets f_H\left( z_L + z_H \right).
\end{align*}
Then, we simply back-propagate through the full recursion process. 

Through deep supervision, the models learns to take any $(z_L,z_H)$ and improve it through a full recursion process, hopefully making $z_H$ closer to the solution. This means that by the design of the deep supervision goal, running a few full recursion processes (even without gradients) is expected to bring us closer to the solution. We propose to run $T-1$ recursion processes without gradient to improve $(z_L,z_H)$ before running one recursion process with backpropagation.

Thus, instead of using the 1-step gradient approximation, we apply a full recursion process containing $n$ evaluations of $f_L$ and 1 evaluation of $f_H$. This removes entirely the need to assume that a fixed-point is reached and the use of the IFT theorem with 1-step gradient approximation. Yet, we can still leverage multiple backpropagation-free recursion processes to improve $(z_L,z_H)$. With this approach, we obtain a massive boost in generalization on Sudoku-Extreme (improving TRM from 56.5\% to 87.4\%; see Table 1).

\subsection{Simpler reinterpretation of $z_H$ and $z_L$} 

HRM is interpreted as doing hierarchical reasoning over two latent features of different hierarchies due to arguments from biology. However, one might wonder why use two latent features instead of 1, 3, or more? And do we really need to justify these so-called "hierarchical" features based on biology to make sense of them? We propose a simple non-biological explanation, which is more natural, and directly answers the question of why there are 2 features.

The fact of the matter is: $z_H$ is simply the current (embedded) solution. The embedding is reversed by applying the output head and rounding to the nearest token using the argmax operation. On the other hand, $z_L$ is a latent feature that does not directly correspond to a solution, but it can be transformed into a solution by applying $z_H \gets f_H(x,z_L,z_H)$. We show an example on Sudoku-Extreme in Figure \ref{fig:sudoku_example} to highlight the fact that $z_H$ does correspond to the solution, but $z_L$ does not. 

Once this is understood, hierarchy is not needed; there is simply an input $x$, a proposed solution $y$ (previously called $z_H$), and a latent reasoning feature $z$ (previously called $z_L$). Given the input question $x$, current solution $y$, and current latent reasoning $z$, the model recursively improves its latent $z$. Then, given the current latent $z$ and the previous solution $y$, the model proposes a new solution $y$ (or stay at the current solution if its already good). 

Although this has no direct influence on the algorithm, this re-interpretation is much simpler and natural. It answers the question about why two features: remembering in context the question $x$, previous reasoning $z$, and previous answer $y$ helps the model iterate on the next reasoning $z$ and then the next answer $y$. If we were not passing the previous reasoning $z$, the model would forget how it got to the previous solution $y$ (since $z$ acts similarly as a chain-of-thought). If we were not passing the previous solution $y$, then the model would forget what solution it had and would be forced to store the solution $y$ within $z$ instead of using it for latent reasoning. Thus, we need both $y$ and $z$ separately, and there is no apparent reason why one would need to split $z$ into multiple features. \newpage

While this is intuitive, we wanted to verify whether using more or less features could be helpful. Results are shown in Table \ref{tab:ablation2}.

\textbf{More features} ($>2$): We tested splitting $z$ into different features by treating each of the $n$ recursions as producing a different $z_i$ for $i=1,...,n$. Then, each $z_i$ is carried across supervision steps. The approach is described in Algorithm \ref{fig:pseudocode_multi}. In doing so, we found performance to drop. This is expected because, as discussed, there is no apparent need for splitting $z$ into multiple parts. It does not have to be hierarchical.

\textbf{Single feature}: Similarly, we tested the idea of taking a single feature by only carrying $z_H$ across supervision steps. The approach is described in Algorithm \ref{fig:pseudocode_single}. In doing so, we found performance to drop. This is expected because, as discussed, it forces the model to store the solution $y$ within $z$.

Thus, we explored using more or less latent variables on Sudoku-Extreme, but found that having only $y$ and $z$ lead to better test accuracy in addition to being the simplest more natural approach.

\begin{table}[ht]
    \caption{TRM on Sudoku-Extreme comparing \% Test accuracy when using more or less latent features}
    \centering
    \begin{tabular}{|l|c|c|c|}
    \hline
     Method & \# of features & Acc (\%) \\
    \hline
TRM $y,z$ (Ours) & 2 & 87.4 \\
TRM multi-scale $z$ & $n+1=7$ & 77.6 \\
TRM single $z$ & 1 & 71.9 \\
\hline
\end{tabular}
\label{tab:ablation2}
\end{table}

\subsection{Single network}

HRM uses two networks, one applied frequently as a \emph{low-level} module $f_H$ and one applied rarely as an \emph{high-level} module ($f_H$). This requires twice the number of parameters compared to regular supervised learning with a single network. 

As mentioned previously, while $f_L$ iterates on the latent reasoning feature $z$ ($z_L$ in HRM), the goal of $f_H$ is to update the solution $y$ ($z_H$ in HRM) given the latent reasoning and current solution. Importantly, since $z\gets f_L(x+y+z)$ contains $x$ but $y\gets f_H(y+z)$ does not contains $x$, the task to achieve (iterating on $z$ versus using $z$ to update $y$) is directly specified by the inclusion or lack of $x$ in the inputs. Thus, we considered the possibility that both networks could be replaced by a single network doing both tasks. In doing so, we obtain better generalization on Sudoku-Extreme (improving TRM from 82.4\% to 87.4\%; see Table 1) while reducing the number of parameters by half. It turns out that a single network is enough.

\subsection{Less is more}

We attempted to increase capacity by increasing the number of layers in order to scale the model. Surprisingly, we found that adding layers decreased generalization due to overfitting. In doing the opposite, decreasing the number of layers while scaling the number of recursions ($n$) proportionally (to keep the amount of compute and emulated depth approximately the same), we found that using 2 layers (instead of 4 layers) maximized generalization. In doing so, we obtain better generalization on Sudoku-Extreme (improving TRM from 79.5\% to 87.4\%; see Table 1) while reducing the number of parameters by half (again).

It is quite surprising that smaller networks are better, but 2 layers seems to be the optimal choice. \citet{bai2024fixed} also observed optimal performance for 2-layers in the context of deep equilibrium diffusion models; however, they had similar performance to the bigger networks, while we instead observe better performance with 2 layers. This may appear unusual, as with modern neural networks, generalization tends to directly correlate with model sizes. However, when data is too scarce and model size is large, there can be an overfitting penalty \citep{kaplan2020scaling}. This is likely an indication that there is too little data. Thus, using tiny networks with deep recursion and deep supervision appears to allow us to bypass a lot of the overfitting. 

\subsection{attention-free architecture for tasks with small fixed context length}

Self-attention is particularly good for long-context lengths when $L \gg D$ since it only requires a matrix of $[D,3D]$ parameters, even though it can account for the whole sequence. However, when focusing on tasks where $L \leq D$, a linear layer is cheap, requiring only a matrix of $[L,L]$ parameters. Taking inspiration from the MLP-Mixer \citep{tolstikhin2021mlp}, we can replace the self-attention layer with a multilayer perceptron (MLP) applied on the sequence length. Using an MLP instead of self-attention, we obtain better generalization on Sudoku-Extreme (improving from 74.7\% to 87.4\%; see Table \ref{tab:ablation}). This worked well on Sudoku 9x9 grids, given the small and fixed context length; however, we found this architecture to be suboptimal for tasks with large context length, such as Maze-Hard and ARC-AGI (both using 30x30 grids). We show results with and without self-attention for all experiments.

\subsection{No additional forward pass needed with ACT}

As previously mentioned, the implementation of ACT in HRM through Q-learning requires two forward passes, which slows down training. We propose a simple solution, which is to get rid of the continue loss (from the Q-learning) and only learn a halting probability through a Binary-Cross-Entropy loss of having reached the correct solution. By removing the continue loss, we remove the need for the expensive second forward pass, while still being able to determine when to halt with relatively good accuracy. We found no significant difference in generalization from this change (going from 86.1\% to 87.4\%; see Table \ref{tab:ablation}).

\subsection{Exponential Moving Average (EMA)}

On small data (such as Sudoku-Extreme and Maze-Hard), HRM tends to overfit quickly and then diverge. To reduce this problem and improves stability, we integrate Exponential Moving Average (EMA) of the weights, a common technique in GANs and diffusion models to improve stability \citep{brock2018large, song2020improved}. We find that it prevents sharp collapse and leads to higher generalization (going from 79.9\% to 87.4\%; see Table \ref{tab:ablation}).

\subsection{Optimal the number of recursions}

We experimented with different number of recursions by varying $T$ and $n$ and found that $T=3,n=3$ (equivalent to 48 recursions) in HRM and $T=3,n=6$ in TRM (equivalent to 42 recursions) to lead to optimal generalization on Sudoku-Extreme. More recursions could be helpful for harder problems (we have not tested it, given our limited resources); however, increasing either $T$ or $n$ incurs massive slowdowns. We show results at different $n$ and $T$ for HRM and TRM in Table \ref{tab:ablation_short}. Note that TRM requires backpropagation through a full recursion process, thus increasing $n$ too much leads to Out Of Memory (OOM) errors. However, this memory cost is well worth its price in gold.

\begin{table}[ht]
    \caption{\% Test accuracy on Sudoku-Extreme dataset. HRM versus TRM matched at a similar effective depth per supervision step $(T(n+1)n_{layers})$}
    \centering
    \begin{tabular}{|ll||cc||cc|}
    \hline
      & & \multicolumn{2}{|c||}{HRM} & \multicolumn{2}{|c|}{TRM} \\
      & & \multicolumn{2}{|c||}{$n=k$, 4 layers} & \multicolumn{2}{|c|}{$n=2k$, 2 layers} \\
      \hline
     $k$ & $T$ & Depth & Acc (\%) & Depth & Acc (\%) \\
    \hline
    1 & 1 & 9 & 46.4 & 7 & 63.2 \\ 
    2 & 2 & 24 & 55.0 & 20 & 81.9 \\ 
    3 & 3 & 48 & 61.6 & 42 & 87.4 \\ 
    4 & 4 & 80 & 59.5 & 72 & 84.2 \\ 
    6 & 3 & 84 & 62.3 & 78 & OOM \\ 
    3 & 6 & 96 & 58.8 & 84 & 85.8 \\ 
    6 & 6 & 168 & 57.5 & 156 & OOM \\ 
    \hline
\end{tabular}
\label{tab:ablation_short}
\end{table}

In the following section, we show our main results on multiple datasets comparing HRM, TRM, and LLMs.

\section{Results}

Following \citet{wang2025hierarchical}, we test our approach on the following datasets: Sudoku-Extreme \citep{wang2025hierarchical}, Maze-Hard \citep{wang2025hierarchical}, ARC-AGI-1 \citep{chollet2019measure} and, ARC-AGI-2 \citep{chollet2025arc}. Results are presented in Tables \ref{tab:puzzle} and \ref{tab:arcgagi}. Hyperparameters are detailed in Section \ref{sec:hyper}. Datasets are discussed below.

Sudoku-Extreme consists of extremely difficult Sudoku puzzles \citep{tdoku, palm2018recurrent, sudoku2018} (9x9 grid), for which only 1K training samples are used to test small-sample learning. Testing is done on 423K samples. Maze-Hard consists of 30x30 mazes generated by the procedure by \citet{lehnert2024beyond} whose shortest path is of length above 110; both the training set and test set include 1000 mazes. 

ARC-AGI-1 and ARC-AGI-2 are geometric puzzles involving monetary prizes. Each puzzle is designed to be easy for a human, yet hard for current AI models. Each puzzle task consists of 2-3 input–output demonstration pairs and 1-2 test inputs to be solved. The final score is computed as the accuracy over all test inputs from two attempts to produce the correct output grid. The maximum grid size is 30x30. ARC-AGI-1 contains 800 tasks, while ARC-AGI-2 contains 1120 tasks. We also augment our data with the 160 tasks from the closely related ConceptARC dataset \citep{moskvichev2023conceptarc}. We provide results on the public evaluation set for both ARC-AGI-1 and ARC-AGI-2. 

While these datasets are small, heavy data-augmentation is used in order to improve generalization. Sudoku-Extreme uses 1000 shuffling (done without breaking the Sudoku rules) augmentations per data example. Maze-Hard uses 8 dihedral transformations per data example. ARC-AGI uses 1000 data augmentations (color permutation, dihedral-group, and translations transformations) per data example. The dihedral-group transformations consist of random 90-degree rotations, horizontal/vertical flips, and reflections.

From the results, we see that TRM without self-attention obtains the best generalization on Sudoku-Extreme (87.4\% test accuracy). Meanwhile, TRM with self-attention generalizes better on the other tasks (probably due to inductive biases and the overcapacity of the MLP on large 30x30 grids). TRM with self-attention obtains 85.3\% accuracy on Maze-Hard, 44.6\% accuracy on ARC-AGI-1, and 7.8\% accuracy on ARC-AGI-2 with 7M parameters. This is significantly higher than the 74.5\%, 40.3\%, and 5.0\% obtained by HRM using 4 times the number of parameters (27M).

\setlength{\tabcolsep}{4pt} 
\begin{table}[ht]
\caption{\% Test accuracy on Puzzle Benchmarks (Sudoku-Extreme and Maze-Hard)}
\centering
\begin{tabular}{|l|c|c|c|}
\hline
 Method & \# Params & Sudoku & Maze \\
\hline
\multicolumn{4}{|c|}{Chain-of-thought, pretrained} \\
\hline
Deepseek R1 & 671B & 0.0 & 0.0 \\
Claude 3.7 8K & ? & 0.0 & 0.0 \\
O3-mini-high & ? & 0.0 & 0.0 \\
\hline
\multicolumn{4}{|c|}{Direct prediction, small-sample training} \\
\hline
Direct pred & 27M & 0.0 & 0.0 \\
HRM & 27M & 55.0 & 74.5 \\
\hdashline
TRM-Att (Ours) & 7M & 74.7 & \highlightgreen{85.3} \\
TRM-MLP (Ours) & 5M/19M\footnotemark & \highlightgreen{87.4} & 0.0 \\
\hline
\end{tabular}
\label{tab:puzzle}
\end{table}
\footnotetext{5M on Sudoku and 19M on Maze}
\setlength{\tabcolsep}{6pt} 

\setlength{\tabcolsep}{4pt} 
\begin{table}[ht]
\caption{\% Test accuracy on ARC-AGI Benchmarks (2 tries)}
\centering
\begin{tabular}{|l|c|c|c|c|c}
\hline
 Method & \# Params & ARC-1 & ARC-2 \\
\hline
\multicolumn{4}{|c|}{Chain-of-thought, pretrained} \\
\hline
Deepseek R1 & 671B & 15.8 & 1.3 \\
Claude 3.7 16K & ? & 28.6 & 0.7 \\
o3-mini-high & ? & 34.5 & 3.0 \\
Gemini 2.5 Pro 32K & ? & 37.0 & 4.9 \\
Grok-4-thinking & 1.7T & 66.7 & 16.0 \\
Bespoke (Grok-4) & 1.7T & \highlightgreen{79.6} & \highlightgreen{29.4} \\
\hline
\multicolumn{4}{|c|}{Direct prediction, small-sample training} \\ 
\hline
Direct pred & 27M & 21.0 & 0.0 \\
HRM & 27M & 40.3 & 5.0 \\
\hdashline
TRM-Att (Ours) & 7M & \highlightgreen{44.6} & \highlightgreen{7.8} \\
TRM-MLP (Ours) & 19M & 29.6 & 2.4 \\
\hline
\end{tabular}
\label{tab:arcgagi}
\end{table}
\setlength{\tabcolsep}{6pt} 

\section{Conclusion}

We propose Tiny Recursion Models (TRM), a simple recursive reasoning approach that achieves strong generalization on hard tasks using a single tiny network recursing on its latent reasoning feature and progressively improving its final answer. Contrary to the Hierarchical Reasoning Model (HRM), TRM requires no fixed-point theorem, no complex biological justifications, and no hierarchy. It significantly reduces the number of parameters by halving the number of layers and replacing the two networks with a single tiny network. It also simplifies the halting process, removing the need for the extra forward pass. Overall, TRM is much simpler than HRM, while achieving better generalization.

While our approach led to better generalization on 4 benchmarks, every choice made is not guaranteed to be optimal on every dataset. For example, we found that replacing the self-attention with an MLP worked extremely well on Sudoku-Extreme (improving test accuracy by 10\%), but poorly on other datasets. Different problem settings may require different architectures or number of parameters. Scaling laws are needed to parametrize these networks optimally. Although we simplified and improved on deep recursion, the question of why recursion helps so much compared to using a larger and deeper network remains to be explained; we suspect it has to do with overfitting, but we have no theory to back this explaination. Not all our ideas made the cut; we briefly discuss some of the failed ideas that we tried but did not work in Section \ref{sec:ideas_failed}. Currently, recursive reasoning models such as HRM and TRM are supervised learning methods rather than generative models. This means that given an input question, they can only provide a single deterministic answer. In many settings, multiple answers exist for a question. Thus, it would be interesting to extend TRM to generative tasks.

\section*{Acknowledgements}

Thank you Emy Gervais for your invaluable support and extra push. This research was enabled in part by computing resources, software, and technical assistance provided by Mila and the Digital Research Alliance of Canada.



\bibliography{paper}
\bibliographystyle{icml2025}

\newpage
\clearpage
\appendix

\section*{Hyper-parameters and setup}\label{sec:hyper}

All models are trained with the AdamW optimizer\citep{loshchilov2017decoupled, kingma2014adam} with $\beta_1=0.9$, $\beta_2=0.95$, small learning rate warm-up (2K iterations), batch-size 768, hidden-size of 512, $N_{sup}=16$ max supervision steps, and stable-max loss \citep{prieto2025grokking} for improved stability. TRM uses an Exponential Moving Average (EMA) of 0.999. HRM uses $n=2,T=2$ with two 4-layers networks, while we use $n=6,T=3$ with one 2-layer network.

For Sudoku-Extreme and Maze-Hard, we train for 60k epochs with learning rate 1e-4 and weight decay 1.0. For ARC-AGI, we train for 100K epochs with learning rate 1e-4 (with 1e-2 learning rate for the embeddings) and weight decay 0.1. The numbers for Deepseek R1, Claude 3.7 8K, O3-mini-high, Direct prediction, and HRM from the Table \ref{tab:puzzle} and \ref{tab:arcgagi} are taken from \citet{wang2025hierarchical}. Both HRM and TRM add an embedding of shape $[0,1,D]$ on Sudoku-Extreme and Maze-Hard to the input. For ARC-AGI, each puzzle (containing 2-3 training examples and 1-2 test examples) at each data-augmentation is given a specific embedding of shape $[0,1,D]$ and, at test-time, the most common answer out of the 1000 data augmentations is given as answer.

Experiments on Sudoku-Extreme were ran with 1 L40S with 40Gb of RAM for generally less than 36 hours. Experiments on Maze-Hard were ran with 4 L40S with 40Gb of RAM for less than 24 hours. Experiments on ARC-AGI were ran for around 3 days with 4 H100 with 80Gb of RAM.

\section*{Ideas that failed}\label{sec:ideas_failed}

In this section, we quickly mention a few ideas that did not work to prevent others from making the same mistake. 

We tried replacing the SwiGLU MLPs by SwiGLU Mixture-of-Experts (MoEs) \citep{shazeer2017outrageously, fedus2022switch}, but we found generalization to decrease massively. MoEs clearly add too much unnecessary capacity, just like increasing the number of layers does.

Instead of back-propagating through the whole $n+1$ recursions, we tried a compromise between HRM 1-step gradient approximation, which back-propagates through the last 2 recursions. We did so by decoupling $n$ from the \emph{number of last recursions} $k$ that we back-propagate through. For example, while $n=6$ requires 7 steps with gradients in TRM, we can use gradients for only the $k=4$ last steps. However, we found that this did not help generalization in any way, and it made the approach more complicated. Back-propagating through the whole $n+1$ recursions makes the most sense and works best.

We tried removing ACT with the option of stopping when the solution is reached, but we found that generalization dropped significantly. This can probably be attributed to the fact that the model is spending too much time on the same data samples rather than focusing on learning on a wide range of data samples.

We tried weight tying the input embedding and output head, but this was too constraining and led to a massive generalization drop.

We tried using TorchDEQ \citep{geng2023torchdeq} to replace the recursion steps by fixed-point iteration as done by Deep Equilibrium Models \citep{bai2019deep}. This would provide a better justification for the 1-step gradient approximation. However, this slowed down training due to the fixed-point iteration and led to worse generalization. This highlights the fact that converging to a fixed-point is not essential.

\clearpage

\section*{Algorithms with different number of latent features}

\begin{figure}[H]
  \centering
    \begin{lstlisting}[language=python]

def latent_recursion(x, z, n=6):
    for i in range(n+1): # latent recursion
        z = net(x, z)
    return z
    
def deep_recursion(x, z, n=6, T=3):
    # recursing T-1 times to improve z (no gradients needed)
    with torch.no_grad():
        for j in range(T-1):
            z = latent_recursion(x, z, n)
    # recursing once to improve z
    z = latent_recursion(x, z, n)
    return z.detach(), output_head(y), Q_head(y)

# Deep Supervision
for x_input, y_true in train_dataloader:
    z = z_init
    for step in range(N_supervision):
        x = input_embedding(x_input)
        z, y_hat, q_hat = deep_recursion(x, z)
        loss = softmax_cross_entropy(y_hat, y_true)
        loss += binary_cross_entropy(q_hat, (y_hat == y_true))
        z = z.detach()
        loss.backward()
        opt.step()
        opt.zero_grad()
        if q[0] > 0: # early-stopping
            break
    \end{lstlisting}
  \caption{Pseudocode of TRM using a single-$z$ with deep supervision training in PyTorch.}
  \label{fig:pseudocode_single}
\end{figure}

\begin{figure}[H]
  \centering
    \begin{lstlisting}[language=python]

def latent_recursion(x, y, z, n=6):
    for i in range(n): # latent recursion
        z[i] = net(x, y, z[0], ... , z[n-1])
    y = net(y, z[0], ... , z[n-1]) # refine output answer
    return y, z
    
def deep_recursion(x, y, z, n=6, T=3):
    # recursing T-1 times to improve y and z (no gradients needed)
    with torch.no_grad():
        for j in range(T-1):
            y, z = latent_recursion(x, y, z, n)
    # recursing once to improve y and z
    y, z = latent_recursion(x, y, z, n)
    return (y.detach(), z.detach()), output_head(y), Q_head(y)

# Deep Supervision
for x_input, y_true in train_dataloader:
    y, z = y_init, z_init
    for step in range(N_supervision):
        x = input_embedding(x_input)
        (y, z), y_hat, q_hat = deep_recursion(x, y, z)
        loss = softmax_cross_entropy(y_hat, y_true)
        loss += binary_cross_entropy(q_hat, (y_hat == y_true))
        loss.backward()
        opt.step()
        opt.zero_grad()
        if q[0] > 0: # early-stopping
            break
    \end{lstlisting}
  \caption{Pseudocode of TRM using multi-scale $z$ with deep supervision training in PyTorch.}
  \label{fig:pseudocode_multi}
\end{figure}

\newpage
\section*{Example on Sudoku-Extreme}

\begin{figure}[H]
\centering
\begin{tikzpicture}[scale=0.38]

  \begin{scope}
    \draw (0, 0) grid (9, 9);
    \draw[very thick, scale=3] (0, 0) grid (3, 3);

    \setcounter{row}{1}
    \setrow { }{}{ }  {}{ }{}  {8}{3}{1}
    \setrow { }{9}{ }  {}{6}{8}  { }{7}{ }
    \setrow { }{}{ }  {3}{ }{5}  { }{}{ }

    \setrow { }{6}{ 8}  {}{ }{}  { }{}{ }
    \setrow { }{}{ }  {}{ }{6}  { }{}{2 }
    \setrow {7 }{4}{ }  {}{ }{}  { }{}{ 3}

    \setrow { }{}{ }  {}{ }{9}  { }{}{4 }
    \setrow {2 }{}{ }  {}{4 }{}  { }{1}{ }
    \setrow {6 }{}{ }  {}{2 }{}  { }{5}{ 7}

    \node[anchor=center] at (4.5, -1.0) {Input $x$};
  \end{scope}

  \begin{scope}[yshift=-11cm]
    \draw (0, 0) grid (9, 9);
    \draw[very thick, scale=3] (0, 0) grid (3, 3);

    \setcounter{row}{1}
    \setrow {5 }{2}{ 6}  {7}{9 }{4}  {8}{3}{1}
    \setrow {3 }{9}{1 }  {2}{6}{8}  { 4}{7}{ 5}
    \setrow {4 }{8}{ 7}  {3}{1 }{5}  {2 }{9}{ 6}

    \setrow {1 }{6}{ 8}  {5}{ 3}{2}  { 7}{4}{9 }
    \setrow {9 }{3}{5 }  {4}{7 }{6}  { 1}{8}{2 }
    \setrow { 7}{4}{2 }  {9}{ 8}{1}  { 5}{6}{3 }

    \setrow { 8}{7}{ 3}  {1}{5 }{9}  { 6}{2}{4 }
    \setrow { 2}{5}{9 }  {6}{ 4}{7}  { 3}{1}{8 }
    \setrow { 6}{1}{4 }  {8}{5 }{3}  { 9}{5}{7 }

    \node[anchor=center] at (4.5, -1.0) {Output $y$};

  \end{scope}

 \begin{scope}[yshift=-22cm]
    \draw (0, 0) grid (9, 9);
    \draw[very thick, scale=3] (0, 0) grid (3, 3);

    \setcounter{row}{1}
    \setrow {5 }{2}{ 6}  {7}{9 }{4}  {8}{3}{1}
    \setrow {3 }{9}{1 }  {2}{6}{8}  { 4}{7}{ 5}
    \setrow {4 }{8}{ 7}  {3}{1 }{5}  {2 }{9}{ 6}

    \setrow {1 }{6}{ 8}  {5}{ 3}{2}  { 7}{4}{9 }
    \setrow {9 }{3}{5 }  {4}{7 }{6}  { 1}{8}{2 }
    \setrow { 7}{4}{2 }  {9}{ 8}{1}  { 5}{6}{3 }

    \setrow { 8}{7}{ 3}  {1}{5 }{9}  { 6}{2}{4 }
    \setrow { 2}{5}{9 }  {6}{ 4}{7}  { 3}{1}{8 }
    \setrow { 6}{1}{4 }  {8}{5 }{3}  { 9}{5}{7 }

    \node[anchor=center] at (4.5, -1.0) {Tokenized $z_H$ (denoted $y$ in TRM)};
  \end{scope}

 \begin{scope}[yshift=-33cm]
    \draw (0, 0) grid (9, 9);
    \draw[very thick, scale=3] (0, 0) grid (3, 3);

    \setcounter{row}{1}
    \setrow {5 }{}{ 5}  {4}{9 }{4}  {}{6}{3}
    \setrow {4 }{}{3 }  {1}{}{}  { 4}{6}{ 5}
    \setrow {4 }{8}{ 4}  {}{3 }{}  {6 }{6}{ 4}

    \setrow {9 }{}{ 6}  {5}{ 3}{}  { 5}{4}{ }
    \setrow { }{3}{5 }  {4}{3 }{}  { 5}{4}{4 }
    \setrow { 6}{}{3 }  {}{ 3}{3}  { 5}{8}{8 }

    \setrow { 3}{3}{ 3}  {6}{5 }{}  { 6}{6}{4 }
    \setrow { 7}{5}{ }  {6}{ }{3}  { 3}{6}{6 }
    \setrow { 4}{3}{4 }  {8}{ }{3}  { 6}{6}{4 }

    \node[anchor=center] at (4.5, -1.0) {Tokenized $z_L$ (denoted $z$ in TRM)};
  \end{scope}

\end{tikzpicture}
  \caption{This Sudoku-Extreme example shows an input, expected output, and the tokenized $z_H$ and $z_L$ (after reversing the embedding and using argmax) for a pretrained model. This highlights the fact that $z_H$ corresponds to the predicted response, while $z_L$ is a latent feature that cannot be decoded to a sensible output unless transformed into $z_H$ by $f_H$.} \label{fig:sudoku_example}
\end{figure}

\end{document}